\documentclass[letterpaper]{article} 
\usepackage{aaai2026}  
\usepackage{times}  
\usepackage{helvet}  
\usepackage{courier}  
\usepackage[hyphens]{url}  
\usepackage{graphicx} 
\usepackage{natbib}  
\usepackage{caption} 
\usepackage{algorithm}
\usepackage{algorithmic}

\usepackage{newfloat}
\usepackage{listings}
\DeclareCaptionStyle{ruled}{labelfont=normalfont,labelsep=colon,strut=off} 
\floatstyle{ruled}
\newfloat{listing}{tb}{lst}{}
\floatname{listing}{Listing}
\usepackage{booktabs}
\usepackage{longtable}
\usepackage{multirow}
\usepackage{array}

\usepackage{amsmath}
\usepackage{amssymb}
\usepackage{subcaption}
\usepackage[utf8]{inputenc}
\usepackage[T1]{fontenc}
\usepackage{times}
\usepackage{pdfpages}
\usepackage[table,xcdraw]{xcolor}

\title{Predict and Resist: Long-Term Accident Anticipation under Sensor Noise}

\author{
 Xingcheng Liu\equalcontrib\textsuperscript{\rm 1},
 Bin Rao\equalcontrib\textsuperscript{\rm 1},
 Yanchen Guan\textsuperscript{\rm1},
 Chengyue Wang\textsuperscript{\rm 1},
 Haicheng Liao\textsuperscript{\rm 1},\\
 Jiaxun Zhang\textsuperscript{\rm 1},
 Chengyu Lin\textsuperscript{\rm 2},
 Meixin Zhu\textsuperscript{\rm 3},
 Zhenning Li\textsuperscript{\rm 1}\thanks{Corresponding Author}}
 \affiliations {
 \textsuperscript{\rm 1}University of Macau\\
 \textsuperscript{\rm 2}Zhejiang University\\
 \textsuperscript{\rm 3}Southeast University\\
 \{yc57431,yc57416,yc37976\}@um.edu.mo,
 chengyue.wang@connect.um.edu.mo,
 \{yc27979,yc47415\}@um.edu.mo,
 linchengyu@zju.edu.cn,
 zhenningli@um.edu.mo} 

\begin{document}

\maketitle

\begin{abstract}
Accident anticipation is essential for proactive and safe autonomous driving, where even a brief advance warning can enable critical evasive actions. However, two key challenges hinder real-world deployment: (1) noisy or degraded sensory inputs from weather, motion blur, or hardware limitations, and (2) the need to issue timely yet reliable predictions that balance early alerts with false-alarm suppression. We propose a unified framework that integrates diffusion-based denoising with a time-aware actor-critic model to address these challenges. The diffusion module reconstructs noise-resilient image and object features through iterative refinement, preserving critical motion and interaction cues under sensor degradation. In parallel, the actor-critic architecture leverages long-horizon temporal reasoning and time-weighted rewards to determine the optimal moment to raise an alert, aligning early detection with reliability. Experiments on three benchmark datasets (DAD, CCD, A3D) demonstrate state-of-the-art accuracy and significant gains in mean time-to-accident, while maintaining robust performance under Gaussian and impulse noise. Qualitative analyses further show that our model produces earlier, more stable, and human-aligned predictions in both routine and highly complex traffic scenarios, highlighting its potential for real-world, safety-critical deployment.

\end{abstract}

\section{Introduction}
\label{sec1}
Traffic accident anticipation—the ability to predict collisions before they occur—represents a critical capability for autonomous driving \cite{ZHANG2025103173}. Unlike traditional perception systems that merely detect accidents after they happen, anticipation enables proactive safety interventions, such as timely braking or evasive maneuvers, potentially preventing collisions entirely \cite{abdel2024matched}. A vehicle that can foresee danger seconds in advance transforms safety from reactive to preventive, which is the ultimate goal of intelligent transportation systems \cite{ref208,ref209}.

However, achieving reliable accident anticipation in real-world driving is profoundly challenging \cite{ref213,ref215} due to two interdependent obstacles:

1) \textbf{Robustness under imperfect perception}. Autonomous vehicles operate with imperfect sensing: rain, glare, dirt, lens damage, and motion blur can obscure critical cues (visual examples are shown in \textbf{Appendix B}). In such noisy conditions, short-term predictions based on single frames become unreliable. Ironically, these are precisely the situations that demand longer temporal reasoning: by accumulating weak signals across time, a model can extract meaningful patterns even when individual frames are corrupted  \cite{croitoru2023diffusion}.

2) \textbf{The problem of “when to warn”}. Most existing approaches focus on frame-level classification or short-horizon prediction \cite{zeng2017agent}: they can indicate if an accident might occur but rarely optimize when to issue an alert \cite{meulemans2023cocoa, pignatelli2023survey}. In safety-critical scenarios, timing is as important as correctness—alerts issued too late are useless, while those issued too early or too often erode trust and can even induce unsafe reactions. This is fundamentally a long-horizon credit assignment problem: the model must identify subtle early cues, maintain temporal reasoning, and determine the optimal moment to act.

\begin{figure}
\centering
\includegraphics[width=0.48\textwidth]{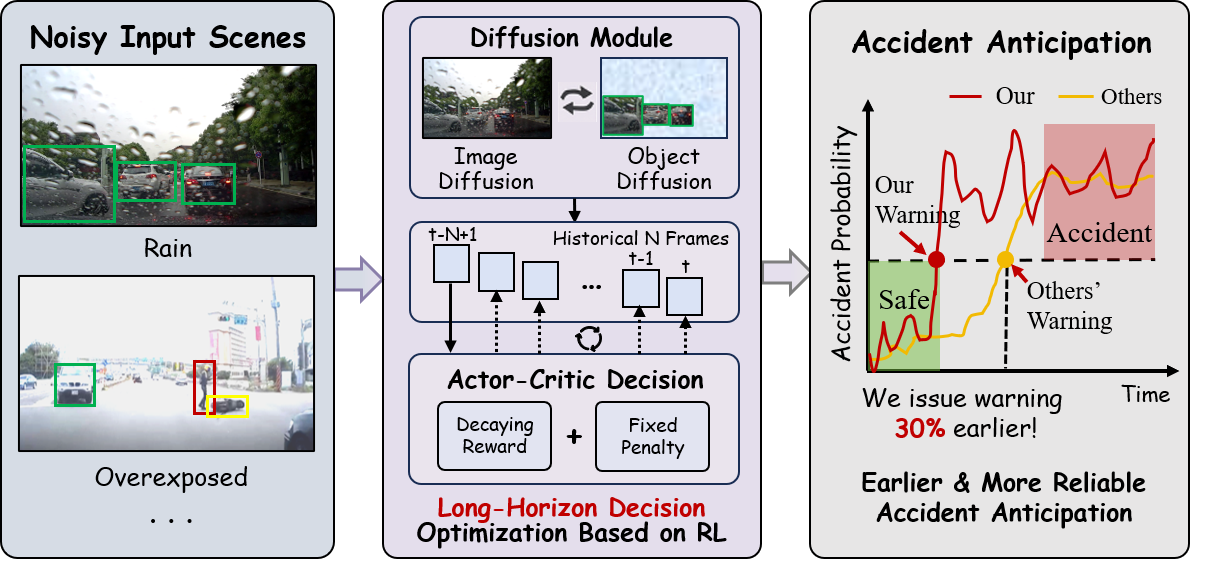}
\caption{Overview illustration of our framework. The figure highlights the integration of the diffusion module and reinforcement learning for processing noisy input scenes, leading to earlier and more reliable accident anticipation.}
\label{tab:Fig. 0.5}
\end{figure}

Crucially, these challenges amplify each other. As shown in Fig.\ref{tab:Fig. 0.5}, sensor degradation increases uncertainty in immediate observations, which in turn magnifies the need for long-horizon temporal reasoning to stabilize predictions. Conversely, a model without effective temporal credit assignment cannot leverage redundancy across frames to overcome noisy perception. A truly deployable anticipation system must therefore treat timing and robustness as a single coupled problem.

To address these challenges, we re-frame accident anticipation as a sequential decision-making problem under uncertainty. Instead of simply classifying frames, our model learns when to warn for maximum safety utility, leveraging an actor-critic reinforcement learning framework for long-horizon credit assignment \cite{sutton2018reinforcement}. To ensure robustness under realistic conditions, we introduce a dual-level diffusion-based denoising module that reconstructs noise-resilient features at both the image and object levels, allowing the system to preserve essential temporal cues even in degraded visual conditions \cite{ho2020denoising, song2021scorebased}. Our dual-level diffusion module acts as a probabilistic feature stabilizer, conceptually similar to Bayesian evidence accumulation. By iteratively refining noisy inputs into structurally faithful and temporally coherent representations, it reduces jitter and spurious activation. This allows the actor-critic module to observe a smoother evolution of risk cues, preventing credit dilution from noisy frames and enabling more effective long-horizon reward assignment. This unified design enables early, reliable, and noise-resilient accident anticipation, bridging the gap between algorithmic capability and real-world deployment.

In summary, the contributions of this paper are as follows:
\begin{itemize}

\item We formulate accident anticipation as a long-horizon credit assignment problem, optimizing not only prediction correctness but also the optimal timing of alerts through an actor-critic framework.

\item We design image-level and object-level diffusion modules to reconstruct robust features under sensor noise, enabling the model to retain critical temporal cues and sustain performance in real-world noisy conditions.

\item Across three benchmark datasets and their noise-augmented variants, our framework achieves state-of-the-art performance in both Average Precision (AP) and mean Time-to-Accident (mTTA), showing that joint long-horizon reasoning and noise-aware feature enhancement yields earlier and more stable warnings than conventional frame-level approaches.

\end{itemize}

\section{Related Work}
Accident anticipation has become a core research problem in autonomous driving because foresight is essential for preventing collisions rather than merely reacting to them. Early studies relied on rule-based heuristics and statistical models, which captured simple patterns but failed to generalize to the complex, multi-agent interactions of real-world traffic \cite{grant2018back}.

The introduction of deep learning has driven a transition to vision-based anticipation, leveraging dashcam or onboard camera video as a rich source of spatial and temporal risk cues \cite{ref22}. The evolution of methods reflects a progression in how the community has tried to model risk:
\begin{itemize}

\item \textbf{From frame-level perception to temporal reasoning.}
Initial deep models relied on CNNs to extract scene appearance and detect static risk cues \cite{ref4,ref9}. While they improved over hand-crafted features, these approaches were myopic, often missing early signals of events that develop gradually. To address this, sequential architectures such as RNNs, LSTMs, and GRUs were introduced \cite{ref10,ref34,ref39}, enabling the capture of risk evolution over time.

\item \textbf{From isolated objects to interaction-aware modeling.}
Anticipating accidents requires understanding how vehicles, pedestrians, and cyclists interact. Graph Neural Networks (GNNs) explicitly encode multi-agent relationships \cite{ref18,ref26,ref36,ref42}, while transformer-based models exploit global attention to capture long-range dependencies and subtle interaction cues \cite{ref11,ref38}. These approaches move beyond simply seeing the scene to reasoning about its dynamics.

\item \textbf{Addressing rarity and interoperability.} True accident events are rare, creating long-tail data challenges. Generative models such as GANs and VAEs synthesize plausible traffic sequences to augment scarce critical scenarios \cite{bao2021drive,ref41}. In parallel, attention mechanisms \cite{ref17,ref18,ref32} and semantic parsing \cite{ref101} improve interpretability by focusing on salient agents and regions, allowing models to highlight the cues most indicative of future risk.

\end{itemize}

Despite this progress, existing methods remain limited in two critical aspects for real-world deployment.
First, they assume clean visual input, yet real-world sensors frequently face rain, glare, blur, or missing pixels, which can obscure subtle pre-accident cues and destabilize short-term predictions.
Second, most models optimize for accident classification rather than the timing of warnings, leaving systems prone to delayed alerts or excessive false positives, both problematic in safety-critical applications.

These gaps highlight the need for methods that combine noise-resilient perception with long-horizon temporal reasoning, enabling accident anticipation that is both early and reliable under real-world conditions.

\section{Methodology}
\subsection{Problem Formulation}
\label{subsec1}

We frame \textit{traffic accident anticipation} as a sequential risk forecasting problem, where the goal is to estimate accident probabilities over time and issue an \textit{early warning} for timely intervention.

Let a video $V = \{V_t\}_{t=1}^T$ consist of frames $V_t$ at time step $t$. A learnable function $f_\theta$ predicts frame-wise accident probabilities $\mathbf{P} = \{p_t\}_{t=1}^T$ as:
\begin{equation}
p_t = f_\theta(V_{1:t}), \quad \text{for } t = 1, \dots, T
\end{equation}
where $V_{1:t}$ is the frame sequence up to time $t$ and $f_\theta$ is parameterized by $\theta$.

To evaluate temporal performance, we define \textit{Time-to-Accident (TTA)} as the interval between the model's first confident prediction and the ground-truth accident frame:
\begin{equation}
\small
\Delta t = \tau - t_o \quad \text{where} \quad t_o = \min\{t \in \{1,\dots,T\} \mid p_t \geq p_{th}\}
\end{equation}
where $\tau$ is the accident frame index (or 0 for negative sequences) and $p_{\text{th}}$ is a decision threshold. A sequence is classified as \textit{accident-positive} if $p_t \geq p_{\text{th}}$ and $\tau > 0$; otherwise, it is \textit{accident-negative}. The model is trained to \textit{optimize} both accuracy and anticipation: discriminating accident-positive/negative sequences while maximizing $\Delta t$ for early warnings with minimal false alarms.



\subsection{Model Framework}
\label{subsec2}

\begin{figure*} [htbp]
\centering
\includegraphics[width=1\textwidth]{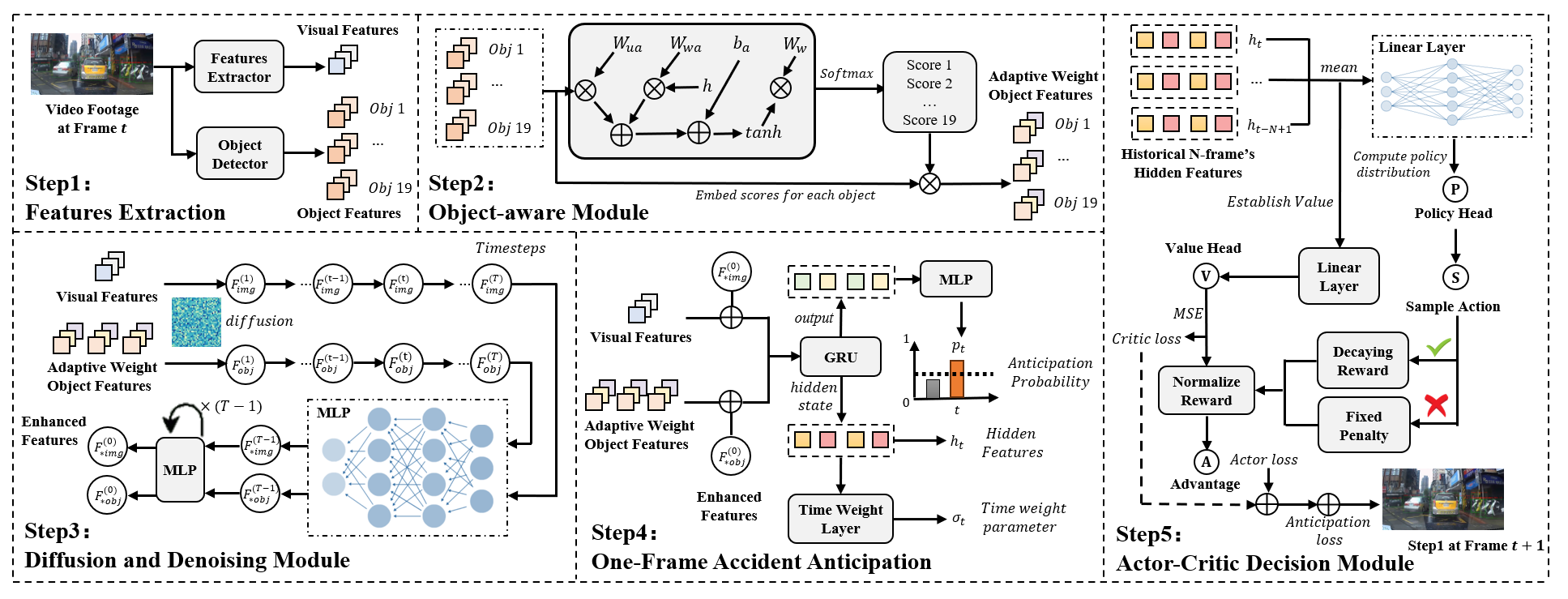}
\caption{Overview of the proposed framework. Input frames are encoded into image and object features, refined by object-aware and diffusion modules. Fused features are processed by a GRU with time-weighted layers to predict $p_t$, while an actor-critic module optimizes long-horizon early warnings.}
\label{tab:Fig. 1}
\end{figure*}

The overall processing pipeline of our framework is illustrated in Fig.~\ref{tab:Fig. 1}. The model consists of five core components: an object detector, a feature extractor, a self-adaptive object-aware module, dual diffusion modules for image- and object-level denoising, and an actor-critic decision module for long-horizon anticipation.

Given an input video sequence, the object detector and feature extractor generate global image features $\mathbf{F}_{img}$ and object-level vectors, refined by the self-adaptive object-aware module to capture dynamic interactions and produce enhanced spatio-temporal representations $\mathbf{\bar{F}}_{obj}$. Both feature sets are denoised via diffusion modules, yielding robust, noise-resilient representations.

The denoised features are fused and processed through a GRU to capture temporal dependencies, generating accident probabilities $p_t$. A time-weighted layer prioritizes critical time segments, while an actor-critic module aggregates long-horizon dependencies, issuing reliable early warnings with minimal false positives. By combining noise-resistant feature extraction with reinforcement-driven temporal reasoning, the framework enables robust accident anticipation under noisy conditions. The algorithm is in the \textbf{Appendix A}. Detailed module descriptions are in subsequent subsections.



\paragraph{Object Detector}  
Each frame is processed by Cascade R-CNN \cite{ref3}, and top-$K$ dynamic agents are encoded into vectors $\mathbf{F}_{obj}$ via VGG-16 \cite{ref31}, preserving appearance and spatial cues.

\paragraph{Feature Extractor}  
Global features $\mathbf{F}_{img}$ are extracted using VGG-16 and an MLP to encode scene context and enhance feature compactness for temporal reasoning.

\paragraph{Self-Adaptive Object-Aware Module}
The object-aware module refines object representations by dynamically attending to the most informative traffic participants based on temporal context and inter-object interactions. 

Given the object features $\mathbf{F}_{obj}$ and the previous hidden state $\mathbf{h}_{t-1}$, attention energies are first computed as:
\begin{equation}
\mathbf{e}_t = \text{tanh}(\mathbf{W}_{wa}\mathbf{h}_{t-1} + \mathbf{W}_{ua}\mathbf{F}_{obj} + \mathbf{b}_a),
\end{equation}
where $\mathbf{W}_{ua}$ and $\mathbf{W}_{wa}$ are learnable projection matrices, and $\mathbf{b}_a$ is a bias vector. These intermediate energies are further transformed as $\mathbf{e}_t' = \mathbf{W}_w\mathbf{e}_t$ and normalized using the softmax function to yield attention weights:$\alpha_t = \text{softmax}(\mathbf{e}_t')$.

The refined object-aware features $\mathbf{\bar{F}}_{obj}$ are obtained by applying the attention weights to the original object features via element-wise multiplication:
\begin{equation}
\mathbf{\bar{F}}_{obj} = \alpha_t \odot \mathbf{F}_{obj}.
\end{equation}
This mechanism adaptively prioritizes high-risk objects while adjusting to evolving scenes and encoding critical interactions temporally to produce robust spatio-temporal representations for reliable accident anticipation.

\paragraph{Diffusion-Based Hierarchical Feature Enhancement}
To improve robustness under noise, we use a diffusion-based module that refines image- and object-level features through noise injection and learned denoising, enabling recovery of meaningful representations from degraded inputs.

1) Adaptive Timestep Sampling. During training, features undergo noise perturbation at a randomly sampled diffusion step $t \sim \mathcal{U}\{0, T-1\}$, where $T$ represents the total steps, improving model generalization and gradient diversity.



2) Variance-Preserving Diffusion. The forward diffusion process perturbs features via a variance-stable Markov chain, ensuring consistent noise scaling across steps.
\begin{align}
\mathbf{F}_{img}^{noisy} &= \sqrt{\bar{\alpha}_t}\mathbf{F}_{img} + \sqrt{1-\bar{\alpha}_t}\epsilon \\
\epsilon &\sim \mathcal{N}(0, \mathbf{I}) \\
\bar{\alpha}_t &= \prod_{s=1}^t \alpha_s \\
\alpha_t &= 1 - \beta_t \\
\beta_t &= \beta_{start} + \frac{t}{T}(\beta_{end} - \beta_{start})
\end{align}
with a linear schedule from $\beta_{start}=0.001$ to $\beta_{end}=0.02$, gradually adding noise while preserving feature variance for smooth transitions between clean and noisy states.

3) Denoising Network Architecture. The denoising network $p_\theta$ refines the noisy image features $\mathbf{F}_{img}^{noisy}$ at each diffusion step $t$ via a lightweight feedforward transformation:
\begin{equation}
p_\theta(\mathbf{F}_{img}^{noisy}, t) = W_2\left(\text{ReLU}(W_1 \mathbf{F}_{img}^{noisy} + b_1)\right) + b_2
\end{equation}
where $W_1, W_2 \in \mathbb{R}^{d \times d}$ and $b_1, b_2 \in \mathbb{R}^{d}$ are learnable parameters. This two-layer structure uses ReLU for non-linearity and preserves input dimensionality, enabling stable feature recovery. Optional step embeddings can integrate timestep information, improving feature reconstruction across varying noise levels.

4) Feature Fusion Strategy. To ensure semantic fidelity under noise, we fuse original and denoised features via residual fusion. The enhanced image feature is computed as
\begin{equation}
\mathbf{F}_{img}^{enhanced} = \mathbf{F}_{img} + \lambda \cdot p_\theta(\mathbf{F}_{img}^{noisy}, t), \quad \lambda = 0.15
\end{equation}
where $\mathbf{F}_{img}$ is the raw image feature, $p_\theta(\cdot)$ is the denoiser and $\lambda$ adjusts residual correction. This design preserves semantics, prevents over-amplification of unstable updates, and ensures stable gradient flow via the identity path.


The same residual enhancement is applied to the refined object-aware features $\bar{\mathbf{F}}_{obj}$:


\begin{equation}
\mathbf{F}_{obj}^{enhanced} = \bar{\mathbf{F}}_{obj} + \lambda \cdot p_\theta(\bar{\mathbf{F}}_{obj}^{noisy}, t)
\end{equation} 
ensuring consistent robustness for both global and local modalities under gradual or abrupt input degradations.

The enhanced image and object features are concatenated and passed through a GRU to capture sequential dependencies, producing the fused representation $\mathbf{X}_t$ and hidden state $\mathbf{h}_t$ for frame $t$:
\begin{equation}
    \mathbf{X_t}, \mathbf{h_t} = \text{GRU}(\text{concat}(\mathbf{F}_{img}^{enhanced},\mathbf{F}_{obj}^{enhanced})
\end{equation}
An MLP then predicts the frame-wise accident probability ${p_t} = \text{MLP}(\mathbf{X_t})$, and a time-weight layer computes the temporal weight loss $w_t = \text{fc}(\mathbf{h_t})$ for the anticipation loss. This unified pipeline produces robust, temporally-aware features for accurate and timely accident anticipation.





\paragraph{State History Processing}
To capture short-term temporal dependencies, a rolling buffer stores the latest $W$ hidden states. Let $\mathbf{h}_i \in \mathbb{R}^d$ be the hidden state at step $i$; then at time $t$, the history $\mathbf{H}_t$ is formed by concatenating the past $W$ states (or all if $t < W$):
\begin{equation}
\mathbf{H}_{t} = \begin{cases}
\text{concat}(\{\mathbf{h}_i\}_{i=1}^t) & \text{if } t < W \\
\text{concat}(\{\mathbf{h}_i\}_{i=t-W+1}^t) & \text{otherwise}
\end{cases}
\end{equation}
Summary vector obtained by averaging the buffered states:
$\bar{\mathbf{h}}_t = \text{mean}(\mathbf{H}_t)$, which smooths fluctuations while retaining key context. This compact representation feeds into decision module, balancing temporal context and efficiency.

\paragraph{Policy and Value Estimation}
We adopt an actor-critic framework to model sequential decision-making, where both the policy (actor) and value (critic) functions are conditioned on the aggregated historical state $\bar{\mathbf{h}}_t$. The actor maps $\bar{\mathbf{h}}_t$ to a discrete action distribution via a linear projection followed by softmax:

\begin{equation}
\pi_t = \text{softmax}(\mathbf{W}_p \bar{\mathbf{h}}_t + \mathbf{b}_p)
\end{equation}
where $\mathbf{W}_p \in \mathbb{R}^{A \times d}$, $\mathbf{b}_p \in \mathbb{R}^A$, and $A$ is the action space size. The action $a_t$ is sampled as $a_t \sim \pi_t$, and its log-probability $\log \pi_t(a_t)$ is retained for policy gradient updates.



The critic predicts the expected cumulative reward from the same state using a linear value head:
\begin{equation}
V_t = \mathbf{w}_v^\top \bar{\mathbf{h}}_t + b_v
\end{equation}
where $\mathbf{w}_v \in \mathbb{R}^d$ and $b_v \in \mathbb{R}$. Decoupling actor and critic stabilizes learning by guiding policy improvement with value-based estimation.

\paragraph{Reward Computation}
The reward function balances prediction correctness and temporal urgency, encouraging early and accurate decisions. The agent receives a positive reward for correct predictions, discounted exponentially to favor earlier actions:
\begin{equation}
r_t = \mathbb{I}(a_t = y_t) \cdot e^{-t/\tau} + \mathbb{I}(a_t \neq y_t) \cdot \gamma
\label{eq:reward}
\end{equation}
where $y_t$ is the ground-truth label, $\tau$ controls temporal decay, and $\gamma$ is a fixed negative penalty for incorrect actions (set to $5$ and $-0.5$ in our experiments).


To stabilize training and reduce reward variance across batches, rewards are normalized as:
\begin{equation}
\tilde{r}_t = \frac{r_t - \mu_r}{\sigma_r + \epsilon}
\end{equation}
where $\mu_r$ and $\sigma_r$ are the batch mean and standard deviation, and $\epsilon$ ensures numerical stability.This reward design promotes early, correct predictions while discouraging late or incorrect actions, aligning with the long-horizon, safety-critical nature of accident anticipation.


\begin{table*}[ht]
    \centering
    \small
    \begin{tabular}{lccccccc}
        \toprule
        \multirow{2}[2]{*}{\textbf{Model}} & \multirow{2}[2]{*}{\textbf{Venue}} & \multicolumn{2}{c}{\textbf{DAD}} & \multicolumn{2}{c}{\textbf{CCD}} & \multicolumn{2}{c}{\textbf{A3D}} \\
        \cmidrule(lr){3-4} \cmidrule(lr){5-6} \cmidrule(lr){7-8}
        & &\textbf{AP (\%)}$\uparrow$ & \textbf{mTTA (s)}$\uparrow$ & \textbf{AP (\%)}$\uparrow$ & \textbf{mTTA (s)}$\uparrow$ & \textbf{AP (\%)}$\uparrow$ & \textbf{mTTA (s)}$\uparrow$ \\
        \midrule
        DSA \cite{ref4} & ACCV & 48.1 & 1.34 & 98.7 & 3.08 & 92.3 & 2.95 \\
        ACRA \cite{zeng2017agent} & CVPR & 51.4 & 3.01 & 98.9 & 3.32 & - & - \\
        AdaLEA \cite{ref33} & CVPR & 52.3 & 3.44 & 99.2 & 3.45 & 92.9 & 3.16 \\
        UString \cite{ref17} & TIV & 53.7 & 3.53 & \underline{99.5} & 3.74 & 93.2 & 3.24 \\
        DSTA \cite{ref1} & ACM MM & 52.9 & 3.21 & 99.1 & 3.54 & 93.5 & 2.87 \\
        GSC \cite{ref37} & TIV & 58.2 & 2.76 & 99.3 & 3.58 & 94.9 & 2.62 \\
        AccNet \cite{ref100} & AAP & 60.8 & 3.58 & \underline{99.5} & 
        3.78 & \underline{95.1} & 3.26 \\
        LATTE \cite{ZHANG2025103173} & IF &  \underline{89.7}  &   \underline{4.49}  &  98.8  & \textbf{4.53}  &  92.5  &   \underline{4.52}  \\
        \cellcolor{purple!15}\textbf{Ours} & \cellcolor{purple!15}- & \cellcolor{purple!15}\textbf{91.2} & \cellcolor{purple!15}\textbf{4.59} & \cellcolor{purple!15}\textbf{99.8} & \cellcolor{purple!15}\underline{4.29} & \cellcolor{purple!15}\textbf{95.7} & \cellcolor{purple!15}\textbf{4.60} \\
        \toprule
    \end{tabular}
    \caption{Comparison of model performance in balancing mTTA and AP across three datasets. Best and second-best values are marked in \textbf{bold} and \underline{underline}, respectively. “–” denotes missing data. ↑ indicates that higher values are better.}
\label{tab:table1}
\end{table*}

\subsection{Training Loss}
\label{subsec1}
The model is trained with a joint objective that integrates \textit{anticipation loss} for supervised prediction and \textit{actor-critic losses} for sequential decision-making, balancing accuracy and early warning capability.

\paragraph{Anticipation Loss}
The supervised anticipation loss $\mathcal{L}_{an}$ combines positive and negative sample terms. For positives, a temporal penalty \(p\) encourages earlier predictions and is defined as $p = -\max\left(0,\displaystyle (t_{\text{accident}} - t_{\text{current}} - 1)/\text{fps} \right)$
and the time weight \(\omega_t\) is computed from the GRU hidden state $h_t$ as $\omega_t = 1 + \sigma(h_t)$, where $\sigma(\cdot)$ is the sigmoid function. The positive loss is:
\begin{equation}
    \mathcal{L}_{\text{pos}} = -\left(\omega_t \cdot \exp(p) \cdot \mathcal{L}_{\text{ce}}\right)
\end{equation}
while the negative loss uses standard cross-entropy scaled by a constant $c$:
$\mathcal{L}_{\text{neg}} = c \mathcal{L}_{\text{ce}}$.

The overall supervised anticipation loss $\mathcal{L}_{an}$ averages over time and samples:
\begin{equation}
    \mathcal{L}_{an} = \mathbb{E}\left[\sum_{i=1}^{T} \left( \alpha_i \cdot \mathcal{L}_{\text{pos}}(i) + (1 - \alpha_i) \cdot \mathcal{L}_{\text{neg}}(i) \right)\right]
\end{equation}


\paragraph{Actor-Critic Losses}
To encourage early and reliable anticipation, we adopt an actor-critic formulation that combines policy and value learning. The \textit{policy loss} guides the actor to favor actions with higher advantages while promoting exploration via entropy regularization:

\begin{equation}
\mathcal{L}_{actor} = -\mathbb{E}[\log \pi_t(a_t) \cdot A_t] - \lambda_e \mathcal{H}(\pi_t)
\end{equation}
where $A_t = \tilde{r}_t - V_t$ is the advantage and $\lambda_e = 0.1$ controls the entropy weight.

The \textit{value loss} encourages the critic to accurately estimate the expected return:
\begin{equation}
\mathcal{L}_{critic} = \frac{1}{2}(\tilde{r}_t - V_t)^2
\end{equation}
with $\tilde{r}_t$ denoting the normalized reward.

Finally, the complete training objective combines the supervised anticipation loss with the actor-critic components:
\begin{equation}
    \mathcal{L}_{total} = \mathcal{L}_{an} + \alpha(\mathcal{L}_{actor} + \beta\mathcal{L}_{critic})
\end{equation}
where $\alpha = \beta = 0.5$ in our implementation.This unified objective balances supervised accuracy with reinforcement-guided timing, driving the model to deliver both \textit{early} and \textit{reliable} accident anticipation.

\section{EXPERIMENT}
\label{4}

\subsection{Experiment Setup}
We evaluate our model on three benchmark datasets covering diverse real-world traffic accidents:

\begin{itemize}
    \item \textbf{Dashcam Accident Dataset (DAD)} \cite{ref4}: 620 accident and 1,130 normal clips (5s@20fps, 100 frames), capturing various urban collisions.
    
    \item \textbf{Car Crash Dataset (CCD)} \cite{ref42}: 1,500 accident and 3,000 normal clips (5s@10fps, 50 frames), with detailed metadata and rich accident diversity.
    
    \item \textbf{AnAn Accident Detection (A3D) Dataset} \cite{ref1}: 1,087 accident and 114 normal clips (5s@20fps, 100 frames), complementing DAD with different urban contexts.
\end{itemize}


To evaluate robustness, we introduce Gaussian and impulse noise at varying levels to simulate sensor degradation, assessing the model’s ability to anticipate accidents accurately under realistic noise conditions.

\subsection{Evaluation Metrics}

$
$
We evaluate the model on two aspects: \textit{accuracy} and \textit{timeliness}, reflecting reliability and early-warning capability. 

\textbf{Accuracy.} We report precision ($P$), recall ($R$), and Average Precision (AP), 
$
AP = \int P(R)\, dR,
$
which measures detection performance across thresholds, with higher AP indicating more reliable recognition.


\textbf{Timeliness.} We measure early-warning performance with Time-to-Accident (TTA), interval between the first confident prediction and the actual accident. Mean TTA (mTTA) is averaged across thresholds as $\text{mTTA} = \int_0^1 \text{TTA}_ a\, da$
where higher values indicate stronger early-warning capability.

\subsection{Implementation Details}


The framework is implemented in PyTorch 2.0 and trained for 30 epochs on an NVIDIA RTX 3050 (batch size 10) using Adam (initial LR $3 \times 10^{-4}$) with a ReduceLROnPlateau scheduler. Each frame includes up to 19 objects with 4096-D features from a VGG-16 backbone. A 256-unit GRU models temporal dynamics for efficient accident anticipation.



\subsection{Evaluation Results}
\noindent\textbf{Compare with SOTA Baselines.} 
Table~1 shows that our model outperforms all baselines in both accuracy (AP) and timeliness (mTTA). On DAD, it achieves 91.2\% AP and 4.59\,s mTTA, reflecting both higher accuracy and earlier anticipation. On CCD and A3D, it delivers consistent AP gains (+0.3\% and 0.6\% than best SOTA) and larger mTTA balance improvements, confirming the advantage of long-horizon temporal reasoning for early warnings.

\begin{table}[ht]
    \centering
    \small
    \begin{tabular}{lcccc}
        \toprule
        \multirow{2}[2]{*}{\textbf{$\sigma$}} & \multicolumn{2}{c}{\textbf{CCD}} & \multicolumn{2}{c}{\textbf{A3D}} \\
        \cmidrule(lr){2-3} \cmidrule(lr){4-5} 
        & \textbf{AP (\%)}$\uparrow$ & \textbf{mTTA (s)}$\uparrow$ & \textbf{AP (\%)}$\uparrow$ & \textbf{mTTA (s)}$\uparrow$ \\
        \hline
        \cellcolor{purple!15}\textbf{Original} & \cellcolor{purple!15}\textbf{99.8} & \cellcolor{purple!15}4.29 & \cellcolor{purple!15}\textbf{95.7} & \cellcolor{purple!15}\textbf{4.60} \\
        0.5 & 99.6 & 4.00 & 94.3 & 4.10 \\
        1.0 & 99.6 & 4.04 & 95.3 & 4.23 \\
        5.0 & 99.6 & \textbf{4.35} & 95.2 & 3.98 \\
        10.0 & 98.0 & 3.43 & 92.9 & 3.97 \\
        20.0 & 91.6 & 3.05 & 92.9 & 3.96 \\
        \bottomrule
    \end{tabular}
    \caption{Comparison of Gaussian noise levels and their impact on mTTA and AP for CCD and A3D. “Original” denotes baseline performance without added noise.}
\label{tab:table3}
\end{table}

\begin{table}[ht]
    \centering
    \small
    \begin{tabular}{lcccc}
        \toprule
        \multirow{2}[2]{*}{\textbf{Percents}} & \multicolumn{2}{c}{\textbf{CCD}} & \multicolumn{2}{c}{\textbf{A3D}} \\
        \cmidrule(lr){2-3} \cmidrule(lr){4-5}
        & \textbf{AP (\%)}$\uparrow$ & \textbf{mTTA (s)}$\uparrow$ & \textbf{AP (\%)}$\uparrow$ & \textbf{mTTA (s)}$\uparrow$ \\
        \hline
        \cellcolor{purple!15}\textbf{Original} & \cellcolor{purple!15}\textbf{99.8} & \cellcolor{purple!15}4.29 & \cellcolor{purple!15}\textbf{95.7} & \cellcolor{purple!15}\textbf{4.60} \\
        10\% & 99.5 & \textbf{4.54} & 95.3 & 4.06 \\
        20\% & 99.6 & 4.33 & 95.7 & 4.37 \\
        30\% & 99.2 & 4.22 & 93.1 & 3.60 \\
        50\% & 98.0 & 3.38 & 91.6 & 3.79 \\
        \bottomrule
    \end{tabular}
    \caption{Comparison of pulse noise levels and their effect on mTTA and AP for CCD and A3D. “Original” indicates baseline performance without added noise.}
\label{tab:table3.5}
\end{table}

\noindent\textbf{Robustness to Sensor Noise.} 
Tables~\ref{tab:table3} and \ref{tab:table3.5} evaluate the model under Gaussian and impulse noise, simulating real-world sensor degradations. Our approach maintains high AP and mTTA even under moderate corruption (e.g., $\sigma=5.0$ for Gaussian noise yields 99.6\% AP on CCD and 95.2\% on A3D). Performance degrades gradually at extreme noise levels but remains meaningful, demonstrating the effectiveness of the diffusion-based denoising module.

Impulse noise shows a similar trend: up to 20\% pixel corruption, the model preserves near-baseline performance, and even at 50\%, it produces usable outputs. This resilience confirms that our framework can sustain reliable accident anticipation under adverse sensing conditions, a critical requirement for real-world deployment. Experiments on robustness of long-horizon credit assignment are in \textbf{Appendix D}.

\subsection{Ablation Studies}
\noindent\textbf{Ablation Study for Core Components.} 
Table~\ref{tab:table4} and \ref{tab:table5} presents an ablation study on the CCD dataset to quantify the contribution of each core component, including the image and object diffusion modules, the self-adaptive object-aware module, the time-weight layer, and the actor-critic components (anticipation, policy, and value losses).


Removing the anticipation loss causes AP to drop sharply to 33.3\%, confirming its key role in aligning predictions with pre-accident cues. This also highlights that a higher mTTA alone is not necessarily better. Value loss removal significantly affects AP and mTTA, highlighting its role in stabilizing long-horizon decisions. The self-adaptive object-aware module and time-weight layer consistently enhance early-warning performance.

\begin{table}[h]
    \centering
        \small
    \begin{tabular}{l|c|c}
        \toprule
        \textbf{Experiment} & \textbf{AP (\%)}$\uparrow$ & \textbf{mTTA (s)}$\uparrow$ \\
        \hline
        \cellcolor{purple!15}\textbf{Our Full Model} & \cellcolor{purple!15}\textbf{99.8} & \cellcolor{purple!15}4.29 \\
        w/o Object Aware Module & 99.3 & 4.61 \\
        w/o Time Weight Layer & 99.5 & 4.47 \\
        w/o Anticipation Loss & 33.3 & \textbf{5.00} \\
        w/o Policy Gradient Loss & 99.6 & 4.47 \\
        w/o Value Loss & 92.8 & 3.03 \\
        \bottomrule
    \end{tabular}
    \caption{Ablation studies of different modules on CCD dataset. "w/o" denotes removal of a module.}
    \label{tab:table4}
\end{table}

\begin{table*}[ht]
    \centering
    \resizebox{0.98\textwidth}{!}{
        \small
    \begin{tabular}{lcccccccc}
        \toprule
        \multirow{2}[2]{*}{\textbf{$\sigma$}} & \multicolumn{2}{c}{\cellcolor{purple!15}\textbf{Original}} & \multicolumn{2}{c}{\textbf{w/o Image Diffusion}} & \multicolumn{2}{c}{\textbf{w/o Object Diffusion}}& \multicolumn{2}{c}{\textbf{w/o All Diffusion}}\\
        \cmidrule(lr){2-3} \cmidrule(lr){4-5} \cmidrule(lr){6-7} \cmidrule(lr){8-9}
        & \cellcolor{purple!15}\textbf{AP (\%)}$\uparrow$ & \cellcolor{purple!15}\textbf{mTTA (s)}$\uparrow$ & \textbf{AP (\%)}$\uparrow$ & \textbf{mTTA (s)}$\uparrow$ & \textbf{AP (\%)}$\uparrow$ & \textbf{mTTA (s)}$\uparrow$& \textbf{AP (\%)}$\uparrow$ & \textbf{mTTA (s)}$\uparrow$\\
        \midrule
        \textbf{Original} & \cellcolor{purple!15}\textbf{99.8} & \cellcolor{purple!15}{4.29} & {99.6} & {4.50} &
        {99.6} & {4.65} & {99.6} & {4.54} \\
        0.5 & \cellcolor{purple!15}\textbf{99.6} & \cellcolor{purple!15}4.00 & 99.4 & 4.45 & 99.5 & 4.35 & 99.4 & 4.36 \\
        1.0 & \cellcolor{purple!15}\textbf{99.6} & \cellcolor{purple!15}4.04 & 99.5 & 4.38 & 99.5 & 4.64 & 99.4 & 4.11 \\
        5.0 & \cellcolor{purple!15}\textbf{99.6} & \cellcolor{purple!15}4.35 & 99.5 & 4.02 & 99.3 & 4.07 & 99.0 & 4.05\\
        10.0 & \cellcolor{purple!15}98.0 & \cellcolor{purple!15}3.43 & 98.6 & 3.75 & \textbf{98.8}& 3.48 & 98.2 & 3.89 \\
        20.0 & \cellcolor{purple!15}91.6 & \cellcolor{purple!15}3.05 & \textbf{92.8} & 3.03 & 91.4 & 3.06 & 91.0 & 3.19 \\
        \bottomrule
    \end{tabular}
    }
    \caption{Impact of varying Gaussian noise levels on mTTA and AP under core module ablation on the CCD dataset. "Original" (column) denotes baseline without noise; "Original" (row) indicates the full model. Best AP per noise level are in bold.}
\label{tab:table5}
\end{table*}

\noindent\textbf{Impact of diffusion modules.}
We analyze the role of dual diffusion modules on the CCD dataset under Gaussian noise ($\sigma \in \{0.5, 1.0, 5.0, 10.0, 20.0\}$) by comparing the full model with variants removing each module or both.

Table~\ref{tab:table5} shows that under clean and mild noise ($\sigma \le 1$), all models perform well, with the full model reaching 99.8\% AP. At moderate noise ($\sigma=5$), dual diffusion best preserves performance (99.6\% AP), while module removal causes slight drops. Under severe noise ($\sigma \ge 10$), all degrade, and omitting image diffusion sometimes improves AP, suggesting over-denoising may harm heavily corrupted inputs.

\subsection{Reward–Penalty Trade-offs in AP and mTTA}

\begin{table}[h]
    \centering
    \small
    \begin{tabular}{llcc}
        \toprule
        \textbf{Reward $\tau$} &
        \textbf{Penalty $\gamma$} &
        \textbf{AP (\%)}$\uparrow$ &
        \textbf{mTTA (s)}$\uparrow$ \\  
        \midrule
        \cellcolor{purple!15}$\times$1 (5.0) & \cellcolor{purple!15}$\times$1 (-0.5) & \cellcolor{purple!15}95.7 & \cellcolor{purple!15}4.60 \\
        $\times$10 & $\times$1 & 93.6 & 4.77 \\
        $\times$50 & $\times$1 & 92.7 & 4.70 \\
        $\times$0.1 & $\times$1 & \textbf{96.2} & 4.47 \\
        $\times$0.02 & $\times$1 & 95.8 & 4.46 \\
        $\times$1 & $\times$10 & 91.2 & \textbf{4.92} \\
        $\times$1 & $\times$0.1 & 92.1 & 4.71 \\
        \bottomrule
    \end{tabular}
    \caption{Effect of reward and penalty scaling on AP and mTTA on A3D. Reward emphasizes early correct alerts; penalty scales false/missed costs.}
    \label{tab:table3.6}
\end{table}

We analyze how the reward coefficient ($\tau$) for early correct predictions and the penalty factor ($\gamma$) for mispredictions affect AP and mTTA on A3D, as summarized in Table~\ref{tab:table3.6}.

Increasing the reward weight (from $\times$1 to $\times$50) steadily reduces AP (95.7\% $\rightarrow$ 92.7\%) while providing slight mTTA gains, indicating that overemphasizing early predictions encourages premature actions and more false positives. Conversely, reducing the reward ($\times$0.1, $\times$0.02) yields the highest AP (up to 96.2\%) but lowers mTTA, reflecting a more conservative strategy that prioritizes accuracy over anticipation. Penalty adjustments show a complementary trend. Strong penalties ($\times$10) make the model overly cautious, achieving the highest mTTA (4.92s) but the lowest AP (91.2\%), while low penalties ($\times$0.1) promote earlier yet less reliable predictions (AP 92.1\%).

\subsection{Visual Analysis: Long-Horizon vs. Frame-Level Training}

\begin{figure}[!h]
\centering
\includegraphics[width=0.5\textwidth,page=1]{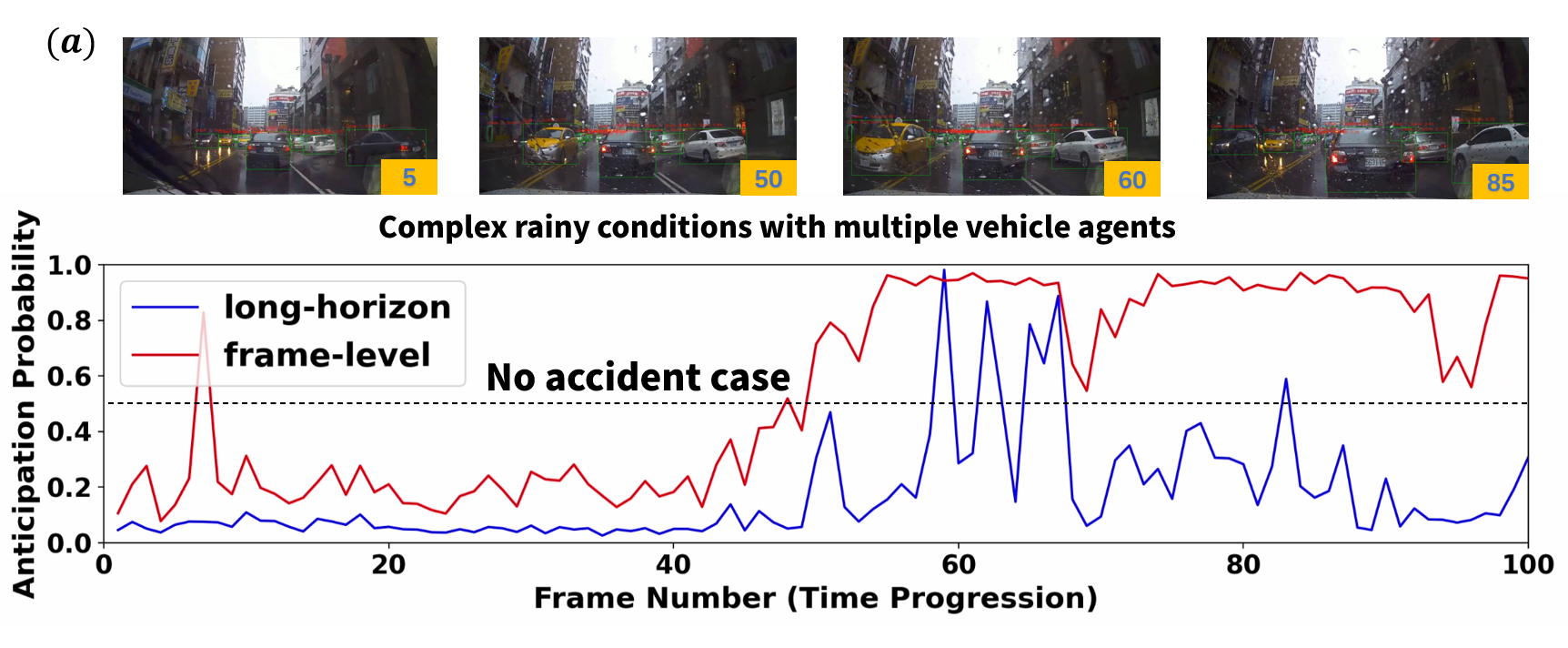}
\includegraphics[width=0.5\textwidth,page=2]{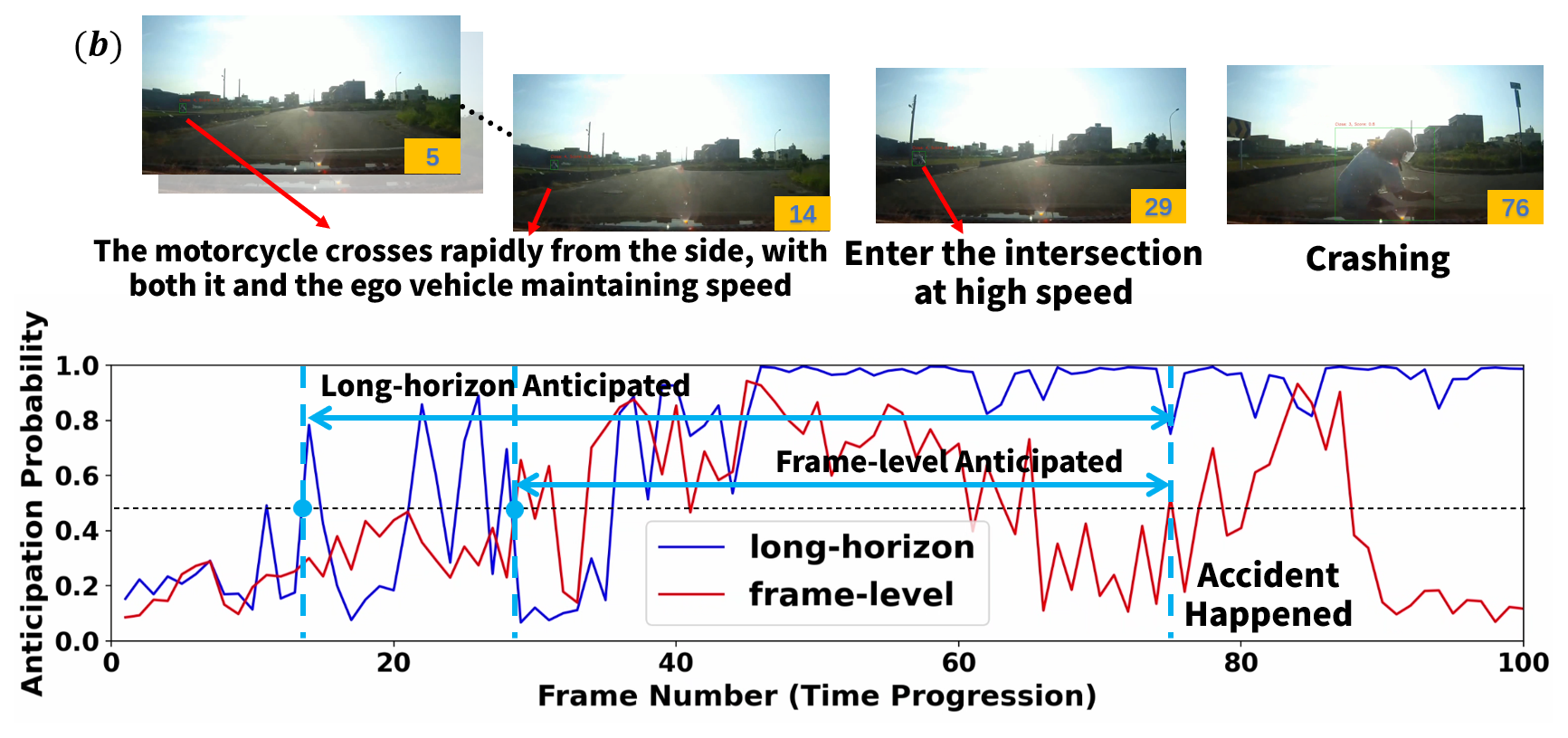} 
\includegraphics[width=0.5\textwidth,page=2]{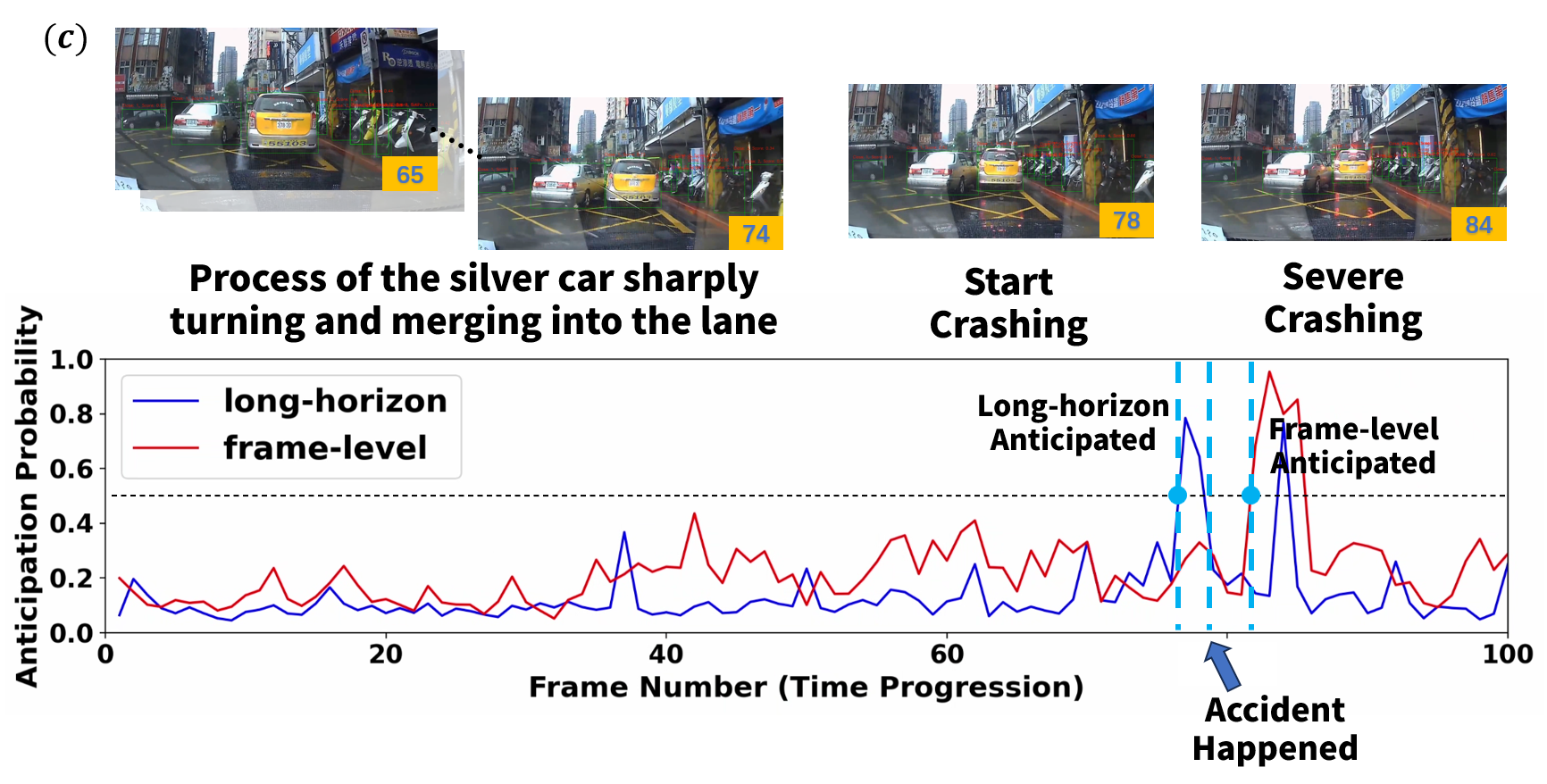} 

\caption{Visualization of long-horizon vs. frame-level anticipation on DAD (threshold 0.5). Scenarios: (a) ambiguous multi-agent rain, (b) predictable collision, (c) sudden complex crash. The long-horizon model offers earlier, more reliable predictions with fewer false positives.}
\label{tab:Fig. 492}
\end{figure}

We compare our long-horizon model (history window=10) with the frame-level baseline (window=0) on three DAD scenarios (Figure~\ref{tab:Fig. 492}). Additional visualizations, window-size comparisons, and inference time are provided in \textbf{Appendices C-G}.

\textbf{(a) Complex multi-agent scenario (False Positive).} In a rainy intersection with multiple interacting vehicles, both models produce false positives. However, the long-horizon model generates shorter and less frequent alarms by integrating temporal evidence from preceding frames, effectively suppressing spurious alerts in noisy environments.

\textbf{(b) Typical collision (True Positive).} 
For a straightforward accident, both models succeed, but the long-horizon model predicts nearly one second earlier by aggregating past frames and capturing subtle cues—such as the motorcycle’s lateral drift without deceleration—enabling timely anticipation before entering the danger zone.

\textbf{(c) Sudden complex crash (Late Prediction).} 
In sudden collisions, although neither model predicts far in advance, the long-horizon model still issues alerts slightly earlier than the frame-level baseline. Even a small lead time can be crucial for timely evasive maneuvers in real driving scenarios.


Overall, these examples show that long-horizon training enables earlier and more reliable warnings with fewer false positives, a critical advantage for safety-critical driving where even brief lead times can greatly reduce risk.

\section{Conclusion}
We proposed a unified framework for traffic accident anticipation that combines diffusion-based denoising with a time-aware actor-critic architecture. This design enhances robustness under noisy sensing and improves the timing of early warnings by leveraging long-horizon temporal reasoning.

Experiments show that our method sustains high accuracy and timeliness on both clean and degraded inputs, while qualitative analysis highlights reduced false positives and earlier predictions compared to frame-level baselines. These results demonstrate that robust, long-horizon anticipation enables safer and more proactive autonomous driving.

\section{Acknowledgments}
This work was supported by the Science and Technology Development Fund of Macau [0122/2024/RIB2, 0215/2024/AGJ, 001/2024/SKL], the Research Services and Knowledge Transfer Office, University of Macau [SRG2023-00037-IOTSC, MYRG-GRG2024-00284-IOTSC], the Shenzhen-Hong Kong-Macau Science and Technology Program Category C [SGDX20230821095159012], the Science and Technology Planning Project of Guangdong [2025A0505010016], National Natural Science Foundation of China [52572354], the State Key Lab of Intelligent Transportation System [2024-B001], and the Jiangsu Provincial Science and Technology Program [BZ2024055].

\bibliography{aaai2026}

@book{sutton2018reinforcement,
  title={Reinforcement Learning: An Introduction},
  author={Sutton, Richard S. and Barto, Andrew G.},
  publisher={MIT Press},
  edition={2nd},
  year={2018}
}

@ARTICLE{croitoru2023diffusion,
author={Croitoru, Florinel-Alin and Hondru, Vlad and Ionescu, Radu Tudor and Shah, Mubarak},
journal={ IEEE Transactions on Pattern Analysis \& Machine Intelligence },
title={{ Diffusion Models in Vision: A Survey }},
year={2023},
volume={45},
number={09},
ISSN={1939-3539},
pages={10850-10869},
doi={10.1109/TPAMI.2023.3261988},
url = {https://doi.ieeecomputersociety.org/10.1109/TPAMI.2023.3261988},
publisher={IEEE Computer Society},
address={Los Alamitos, CA, USA},
month=sep}

@inproceedings{
song2021scorebased,
title={Score-Based Generative Modeling through Stochastic Differential Equations},
author={Yang Song and Jascha Sohl-Dickstein and Diederik P Kingma and Abhishek Kumar and Stefano Ermon and Ben Poole},
booktitle={International Conference on Learning Representations},
year={2021},
url={https://openreview.net/forum?id=PxTIG12RRHS}
}

@inproceedings{ho2020denoising,
author = {Ho, Jonathan and Jain, Ajay and Abbeel, Pieter},
title = {Denoising diffusion probabilistic models},
year = {2020},
isbn = {9781713829546},
publisher = {Curran Associates Inc.},
address = {Red Hook, NY, USA},
booktitle = {Proceedings of the 34th International Conference on Neural Information Processing Systems},
articleno = {574},
numpages = {12},
location = {Vancouver, BC, Canada},
series = {NIPS '20}
}

@inproceedings{bao2021drive,
  title={DRIVE: Deep ReInforced Accident Anticipation with Visual Explanation},
  author={Bao, Wentao and Yu, Qi and Kong, Yu},
  booktitle={2021 IEEE/CVF International Conference on Computer Vision (ICCV)}, 
  title={DRIVE: Deep Reinforced Accident Anticipation with Visual Explanation}, 
  year={2021},
  volume={},
  number={},
  pages={7599-7608},
  doi={10.1109/ICCV48922.2021.00752}
}

@article{abdel2024matched,
  title={A matched case-control analysis of autonomous vs human-driven vehicle accidents},
  author={Abdel-Aty, Mohamed and Ding, Shengxuan},
  journal={Nature Communications},
  volume={15},
  number={1},
  pages={2024--2033},
  year={2024},
  doi={10.1038/s41467-024-48526-4}
}

@inproceedings{meulemans2023cocoa,
author = {Meulemans, Alexander and Schug, Simon and Kobayashi, Seijin and Daw, Nathaniel D. and Wayne, Gregory},
title = {Would I have gotten that reward? long-term credit assignment by counterfactual contribution analysis},
year = {2023},
publisher = {Curran Associates Inc.},
address = {Red Hook, NY, USA},
booktitle = {Proceedings of the 37th International Conference on Neural Information Processing Systems},
articleno = {3006},
numpages = {51},
location = {New Orleans, LA, USA},
series = {NIPS '23}
}

@article{
pignatelli2023survey,
title={A Survey of Temporal Credit Assignment in Deep Reinforcement Learning},
author={Eduardo Pignatelli and Johan Ferret and Matthieu Geist and Thomas Mesnard and Hado van Hasselt and Laura Toni},
journal={Transactions on Machine Learning Research},
issn={2835-8856},
year={2024},
url={https://openreview.net/forum?id=bNtr6SLgZf},
note={Survey Certification}
}

@INPROCEEDINGS{zeng2017agent,
  author={Zeng, Kuo-Hao and Chou, Shih-Han and Chan, Fu-Hsiang and Niebles, Juan Carlos and Sun, Min},
  booktitle={2017 IEEE Conference on Computer Vision and Pattern Recognition (CVPR)}, 
  title={Agent-Centric Risk Assessment: Accident Anticipation and Risky Region Localization}, 
  year={2017},
  volume={},
  number={},
  pages={1330-1338},
  doi={10.1109/CVPR.2017.146}}

@article{ZHANG2025103173,
title = {LATTE: A Real-time Lightweight Attention-based Traffic Accident Anticipation Engine},
author = {Jiaxun Zhang and Yanchen Guan and Chengyue Wang and Haicheng Liao and Guohui Zhang and Zhenning Li},
journal = {Information Fusion},
volume = {122},
pages = {103173},
year = {2025},
issn = {1566-2535},
doi = {https://doi.org/10.1016/j.inffus.2025.103173},
url = {https://www.sciencedirect.com/science/article/pii/S1566253525002465}
}

@article{ref100,
title = {Real-time accident anticipation for autonomous driving through monocular depth-enhanced 3D modeling},
journal = {Accident Analysis \& Prevention},
volume = {207},
pages = {107760},
year = {2024},
issn = {0001-4575},
doi = {10.1016/j.aap.2024.107760},
url = {https://www.sciencedirect.com/science/article/pii/S0001457524003051},
author = {Haicheng Liao and Yongkang Li and Zhenning Li and Zilin Bian and Jaeyoung Lee and Zhiyong Cui and Guohui Zhang and Chengzhong Xu}
}

@article{grant2018back,
  title={Back to the future: What do accident causation models tell us about accident prediction?},
  author={Grant, Eryn and Salmon, Paul M and Stevens, Nicholas J and Goode, Natassia and Read, Gemma J},
  journal={Safety Science},
  volume={104},
  pages={99--109},
  year={2018},
  publisher={Elsevier},
  doi = {10.1016/j.ssci.2017.12.018}
}

@InProceedings{ref4,
author="Chan, Fu-Hsiang
and Chen, Yu-Ting
and Xiang, Yu
and Sun, Min",
editor="Lai, Shang-Hong
and Lepetit, Vincent
and Nishino, Ko
and Sato, Yoichi",
title="Anticipating Accidents in Dashcam Videos",
doi="10.1007/978-3-319-54190-7_9",
booktitle="Computer Vision --  ACCV 2016",
year="2017",
publisher="Springer International Publishing",
address="Cham",
pages="136--153",
isbn="978-3-319-54190-7"
}

@inproceedings{ref42,
author = {Yao, Yu and Xu, Mingze and Choi, Chiho and Crandall, David J. and Atkins, Ella M. and Dariush, Behzad},
title = {Egocentric Vision-based Future Vehicle Localization for Intelligent Driving Assistance Systems},
year = {2019},
publisher = {IEEE Press},
url = {https://doi.org/10.1109/ICRA.2019.8794474},
doi = {10.1109/ICRA.2019.8794474},
booktitle = {2019 International Conference on Robotics and Automation (ICRA)},
pages = {9711–9717},
numpages = {7},
location = {Montreal, QC, Canada}
}

@InProceedings{ref1,
    author = {Bao, Wentao and Yu, Qi and Kong, Yu},
    title  = {Uncertainty-based Traffic Accident Anticipation with Spatio-Temporal Relational Learning},
    doi = {10.1145/3394171.3413827},
    booktitle = {Proceedings of the 28th ACM International Conference on Multimedia (MM ’20)},
    month  = {October},
    year   = {2020}
}

@article{ref22,
title = {Mitigating the impact of outliers in traffic crash analysis: A robust Bayesian regression approach with application to tunnel crash data},
journal = {Accident Analysis \& Prevention},
volume = {185},
pages = {107019},
year = {2023},
issn = {0001-4575},
doi = {10.1016/j.aap.2023.107019},
url = {https://www.sciencedirect.com/science/article/pii/S0001457523000660},
author = {Zhenning Li and Haicheng Liao and Ruru Tang and Guofa Li and Yunjian Li and Chengzhong Xu}
}

@ARTICLE{ref9,
  author={Fang, Jianwu and Qiao, Jiahuan and Bai, Jie and Yu, Hongkai and Xue, Jianru},
  journal={IEEE Transactions on Intelligent Transportation Systems}, 
  title={Traffic Accident Detection via Self-Supervised Consistency Learning in Driving Scenarios}, 
  year={2022},
  volume={23},
  number={7},
  pages={9601-9614},
  doi={10.1109/TITS.2022.3157254}}

@INPROCEEDINGS{ref10,
  author={Fatima, Mishal and Karim Khan, Muhammad Umar and Kyung, Chong-Min},
  booktitle={2020 25th International Conference on Pattern Recognition (ICPR)}, 
  title={Global Feature Aggregation for Accident Anticipation}, 
  year={2021},
  volume={},
  number={},
  pages={2809-2816},
  doi={10.1109/ICPR48806.2021.9412338}}

@inproceedings{ref34,
author = {Takimoto, Yoshiaki and Tanaka, Yusuke and Kurashima, Takeshi and Yamamoto, Shuhei and Okawa, Maya and Toda, Hiroyuki},
title = {Predicting Traffic Accidents with Event Recorder Data},
year = {2019},
isbn = {9781450369640},
publisher = {Association for Computing Machinery},
address = {New York, NY, USA},
url = {https://doi.org/10.1145/3356995.3364535},
doi = {10.1145/3356995.3364535},
booktitle = {Proceedings of the 3rd ACM SIGSPATIAL International Workshop on Prediction of Human Mobility},
pages = {11–14},
numpages = {4},
location = {Chicago, IL, USA},
series = {PredictGIS'19}
}

@INPROCEEDINGS{ref39,
  author={Xue, Ruoyu and Chen, Jingyuan and Fang, Yajun},
  booktitle={2020 5th International Conference on Universal Village (UV)}, 
  title={Real-Time Anomaly Detection and Feature Analysis Based on Time Series for Surveillance Video}, 
  year={2020},
  volume={},
  number={},
  pages={1-7},
  doi={10.1109/UV50937.2020.9426191}}

@ARTICLE{ref18,
  author={Karim, Muhammad Monjurul and Yin, Zhaozheng and Qin, Ruwen},
  journal={IEEE Transactions on Intelligent Vehicles}, 
  title={An Attention-Guided Multistream Feature Fusion Network for Early Localization of Risky Traffic Agents in Driving Videos}, 
  year={2024},
  volume={9},
  number={1},
  pages={1792-1803},
  doi={10.1109/TIV.2023.3275543}}

@inproceedings{ref26,
author = {Liu, Kun and Zhu, Minzhi and Fu, Huiyuan and Ma, Huadong and Chua, Tat-Seng},
title = {Enhancing Anomaly Detection in Surveillance Videos with Transfer Learning from Action Recognition},
year = {2020},
isbn = {9781450379885},
publisher = {Association for Computing Machinery},
address = {New York, NY, USA},
url = {https://doi.org/10.1145/3394171.3416298},
doi = {10.1145/3394171.3416298},
booktitle = {Proceedings of the 28th ACM International Conference on Multimedia},
pages = {4664–4668},
numpages = {5},
location = {Seattle, WA, USA},
series = {MM '20}
}

@InProceedings{ref36,
    author    = {Thakur, Nupur and Gouripeddi, PrasanthSai and Li, Baoxin},
    title     = {Graph(Graph): A Nested Graph-Based Framework for Early Accident Anticipation},
    booktitle = {Proceedings of the IEEE/CVF Winter Conference on Applications of Computer Vision (WACV)},
    month     = {January},
    year      = {2024},
    pages     = {7533-7541},
    doi = {10.1109/WACV57701.2024.00736}
}

@ARTICLE{ref37,
  author={Wang, Tianhang and Chen, Kai and Chen, Guang and Li, Bin and Li, Zhijun and Liu, Zhengfa and Jiang, Changjun},
  journal={IEEE Transactions on Intelligent Vehicles}, 
  title={GSC: A Graph and Spatio-Temporal Continuity Based Framework for Accident Anticipation}, 
  year={2024},
  volume={9},
  number={1},
  pages={2249-2261},
  doi={10.1109/TIV.2023.3257169}}

@ARTICLE{ref41,
  author={Yao, Yu and Wang, Xizi and Xu, Mingze and Pu, Zelin and Wang, Yuchen and Atkins, Ella and Crandall, David J.},
  journal={IEEE Transactions on Pattern Analysis and Machine Intelligence}, 
  title={DoTA: Unsupervised Detection of Traffic Anomaly in Driving Videos}, 
  year={2023},
  volume={45},
  number={1},
  pages={444-459},
  doi={10.1109/TPAMI.2022.3150763}}

@INPROCEEDINGS{ref11,
  author={Feng, Jia-Chang and Hong, Fa-Ting and Zheng, Wei-Shi},
  booktitle={2021 IEEE/CVF Conference on Computer Vision and Pattern Recognition (CVPR)}, 
  title={MIST: Multiple Instance Self-Training Framework for Video Anomaly Detection}, 
  year={2021},
  volume={},
  number={},
  pages={14004-14013},
  doi={10.1109/CVPR46437.2021.01379}}

@inproceedings{ref38,
  title     = {Weakly-Supervised Spatio-Temporal Anomaly Detection in Surveillance Video},
  author    = {Wu, Jie and Zhang, Wei and Li, Guanbin and Wu, Wenhao and Tan, Xiao and Li, Yingying and Ding, Errui and Lin, Liang},
  booktitle = {Proceedings of the Thirtieth International Joint Conference on
               Artificial Intelligence, {IJCAI-21}},
  publisher = {International Joint Conferences on Artificial Intelligence Organization},
  editor    = {Zhi-Hua Zhou},
  pages     = {1172--1178},
  year      = {2021},
  month     = {8},
  note      = {Main Track},
  doi       = {10.24963/ijcai.2021/162},
  url       = {https://doi.org/10.24963/ijcai.2021/162},
}

@article{ref17,
author = {Karim, Muhammad Monjurul and Li, Yu and Qin, Ruwen and Yin, Zhaozheng},
title = {A Dynamic Spatial-Temporal Attention Network for Early Anticipation of Traffic Accidents},
year = {2022},
issue_date = {July 2022},
publisher = {IEEE Press},
volume = {23},
number = {7},
issn = {1524-9050},
url = {https://doi.org/10.1109/TITS.2022.3155613},
doi = {10.1109/TITS.2022.3155613},
journal = {Trans. Intell. Transport. Sys.},
month = jul,
pages = {9590–9600},
numpages = {11}
}

@article{ref32,
title = {Dynamic attention augmented graph network for video accident anticipation},
journal = {Pattern Recognition},
volume = {147},
pages = {110071},
year = {2024},
issn = {0031-3203},
doi = {10.1016/j.patcog.2023.110071},
url = {https://www.sciencedirect.com/science/article/pii/S0031320323007689},
author = {Wenfeng Song and Shuai Li and Tao Chang and Ke Xie and Aimin Hao and Hong Qin}
}

@INPROCEEDINGS{ref3,
  author={Cai, Zhaowei and Vasconcelos, Nuno},
  booktitle={2018 IEEE/CVF Conference on Computer Vision and Pattern Recognition}, 
  title={Cascade R-CNN: Delving Into High Quality Object Detection}, 
  year={2018},
  volume={},
  number={},
  pages={6154-6162},
  doi={10.1109/CVPR.2018.00644}}

@inproceedings{ref31,
  edition = {},
  number = {},
  booktitle = {3rd International Conference on Learning Representations (ICLR 2015)},
  pages = {1-14},
  publisher = {Computational and Biological Learning Society},
  school = {},
  title = {Very deep convolutional networks for large-scale image recognition},
  volume = {},
  author = {Simonyan, K and Zisserman, A},
  editor = {},
  year = {2015},
  organizer = {},
  series = {},
  url = {https://arxiv.org/abs/1409.1556}
}

@InProceedings{ref33,
author = {Suzuki, Tomoyuki and Kataoka, Hirokatsu and Aoki, Yoshimitsu and Satoh, Yutaka},
title = {Anticipating Traffic Accidents With Adaptive Loss and Large-Scale Incident DB},
booktitle = {Proceedings of the IEEE Conference on Computer Vision and Pattern Recognition (CVPR)},
month = {June},
year = {2018},
doi={10.1109/CVPR.2018.00371}
}

@inproceedings{ref101,
author = {Liao, Haicheng and Sun, Haoyu and Shen, Huanming and Wang, Chengyue and Tian, Chunlin and Tam, KaHou and Li, Li and Xu, Chengzhong and Li, Zhenning},
title = {CRASH: Crash Recognition and Anticipation System Harnessing with Context-Aware and Temporal Focus Attentions},
year = {2024},
isbn = {9798400706868},
publisher = {Association for Computing Machinery},
address = {New York, NY, USA},
url = {https://doi.org/10.1145/3664647.3680672},
doi = {10.1145/3664647.3680672},
booktitle = {Proceedings of the 32nd ACM International Conference on Multimedia},
pages = {11041–11050},
numpages = {10},
location = {Melbourne VIC, Australia},
series = {MM '24}
}

@article{ref208,
author = {Fang, Jianwu and Qiao, Jiahuan and Xue, Jianru and Li, Zhengguo},
title = {Vision-Based Traffic Accident Detection and Anticipation: A Survey},
year = {2024},
issue_date = {April 2024},
publisher = {IEEE Press},
volume = {34},
number = {4},
issn = {1051-8215},
url = {https://doi.org/10.1109/TCSVT.2023.3307655},
doi = {10.1109/TCSVT.2023.3307655},
journal = {IEEE Trans. Cir. and Sys. for Video Technol.},
month = apr,
pages = {1983–1999},
numpages = {17}
}

@article{ref209,
title = {Analysis of safety benefits and security concerns from the use of autonomous vehicles: A grouped random parameters bivariate probit approach with heterogeneity in means},
journal = {Analytic Methods in Accident Research},
volume = {28},
pages = {100134},
year = {2020},
issn = {2213-6657},
doi = {10.1016/j.amar.2020.100134},
url = {https://www.sciencedirect.com/science/article/pii/S2213665720300245},
author = {Sheikh Shahriar Ahmed and Sarvani Sonduru Pantangi and Ugur Eker and Grigorios Fountas and Stephen E. Still and Panagiotis Ch. Anastasopoulos}
}

@article{ref213,
title = {Advances, challenges, and future research needs in machine learning-based crash prediction models: A systematic review},
journal = {Accident Analysis \& Prevention},
volume = {194},
pages = {107378},
year = {2024},
issn = {0001-4575},
doi = {10.1016/j.aap.2023.107378},
url = {https://www.sciencedirect.com/science/article/pii/S0001457523004256},
author = {Yasir Ali and Fizza Hussain and Md Mazharul Haque}
}

@inproceedings{ref215,
  title     = {MFTraj: Map-Free, Behavior-Driven Trajectory Prediction for Autonomous Driving},
  author    = {Liao, Haicheng and Li, Zhenning and Wang, Chengyue and Shen, Huanming and Liao, Dongping and Wang, Bonan and Li, Guofa and Xu, Chengzhong},
  booktitle = {Proceedings of the Thirty-Third International Joint Conference on
               Artificial Intelligence, {IJCAI-24}},
  publisher = {International Joint Conferences on Artificial Intelligence Organization},
  editor    = {Kate Larson},
  pages     = {5945--5953},
  year      = {2024},
  month     = {8},
  note      = {Main Track},
  doi       = {10.24963/ijcai.2024/657},
  url       = {https://doi.org/10.24963/ijcai.2024/657},
}


\end{document}